\pdfoutput=1

\documentclass[11pt]{article}

\usepackage[]{EMNLP2023}

\usepackage{times}
\usepackage{latexsym}

\usepackage[T1]{fontenc}

\usepackage[utf8]{inputenc}

\usepackage{microtype}

\usepackage{inconsolata}

\usepackage{algorithm}
\usepackage{algorithmic}

\title{Exploring the traditional NMT model and Large Language Model for chat translation}

\author{
	Jinlong Yang,
        Hengchao Shang,
        Daimeng Wei,
        Jiaxin Guo, \\
	\textbf{
        Zongyao Li,
        Zhanglin Wu,
		Zhiqiang Rao,
		Shaojun Li,
		}\\
	\textbf{
		Yuhao Xie,
		Yuanchang Luo,
        Jiawei Zheng,
        Bin Wei,
		Hao Yang}\\
	Huawei Translation Service Center, Beijing, China\\
	\{yangjinlong7,shanghengchao,weidaimeng,guojiaxin1,lizongyao,\\
	wuzhanglin2,raozhiqiang,lishaojun18,xieyuhao2,luoyuanchang,\\
	zhengjiawei15,weibin29,yanghao30\}@huawei.com \\
}

\begin{document}
\maketitle
\begin{abstract}
This paper describes the submissions of Huawei Translation Services Center(HW-TSC) to  WMT24 chat translation shared task on English$\leftrightarrow$Germany (en-de) bidirection. The experiments involved fine-tuning models using chat data and exploring various strategies, including Minimum Bayesian Risk (MBR) decoding and self-training. The results show significant performance improvements in certain directions, with the MBR self-training method achieving the best results. The Large Language Model also discusses the challenges and potential avenues for further research in the field of chat translation.
\end{abstract}

\section{Introduction}
Neural machine translation (NMT) \citep{sutskever2014sequence, bahdanau2015neural, gehring2017convolutional, wu-etal-2023-improving} has made substantial progress in recent years, largely due to the adoption of the transformer \citep{vaswani2017attention} architecture. NMT has demonstrated promising translation results across various scenarios. However, research in the field of chat translation remains limited, primarily due to the scarcity of chat data. In prior chat-related tasks, we utilized data from related domains, such as spoken dialogue and subtitles, to augment our translation models, but the outcomes were only mediocre.

Like the preceding two chat shared tasks, the WMT24 chat shard task concentrates on translating conversations between consumers and servers in different languages. We participated in the en-de bidirectional translation task. The en-de bidirectional models we submitted to the WMT22 chat task \citep{yang-etal-2022-hw} function as our baseline models, leveraging the deep transformer \citep{dou2018exploiting} architecture. Building on this foundation, we employed the Minimum Bayesian Risk (MBR) strategy to select the optimal translation outcomes, and iterative self-training yielded the best results on the development set.

Beyond traditional NMT models, the emergence of large language model(LLM) has introduced a new paradigm to translation tasks\citep{Wang_Lyu_Ji_Zhang_Yu_Shi_Tu, Moslem_Haque_Way_2023,guo2024novelparadigmboostingtranslation}. Due to its extensive context length and powerful language modeling capabilities, large language models significantly outperform NMT in the translation of lengthy texts and the fluency of translation results. We input the translation output from the NMT model into the LLM as a prompt, allowing the LLM to combine the reference translatio from traditional NMT model to produce an improved translation. However, the comet metric of the LLM’s output did not surpass the optimal results of the NMT model.

Recognizing that chat translation is a context-aware task, we conducted a series of context-aware experiments\citep{wu2024adapting} using LLMs with WMT and IWSLT document data . We fine-tuned the LLM by constructing streamed translations and contextualized translation data, and translated the development set in the same format. Unfortunately, the results were unsatisfactory.

The structure of this paper is as follows: Section~\ref{sec:data} describes our data volume and format for fine-tuning the LLM. The model structure and key methods utilized are presented in Section~\ref{sec:system}. Section~\ref{sec:setting} outlines our experiment setting. Results and analysis are presented in Section~\ref{sec:result}, and we conclude our work in Section~\ref{sec:conclusion}.

\section{Data}
\label{sec:data}
\subsection{Data Size}
All experiments conducted for this task are based on the model developed by our team, as participated in the WMT22 chat shared task. For details on the training data and strategies used for this model, please refer to the system report \citet{yang-etal-2022-hw,wei-etal-2021-hw}. Table~\ref{tab:bi-data} and Table~\ref{tab:doc-data} list all the data used in this experiment. Based on the prior tasks experience, the contribution of out-domain data to the improvement of translation quality is limited. Therefore, we only further optimize our translation model using the data shown in Table~\ref{tab:bi-data}, which consists of historical chat tasks. The data in Table~\ref{tab:doc-data} is used for fine-tuning the LLM, enabling it to translate context-aware texts and validate the impact of paragraph information on dialogue translation quality.

\begin{table}
	\centering
\begin{tabular}{cccc}
\hline
\textbf{24 train} & \textbf{24 valid} & \textbf{22 valid} & \textbf{22 test}\\
\hline
17805 & 2569 & 2109 & 2488 \\
\hline
\end{tabular}
\caption{Chat shared task en-de bilingual data lines used for training}
\label{tab:bi-data}
\end{table}

\begin{table}
	\centering
	\small
	\begin{tabular}{ccc}
		\hline
		\textbf{Dataset}    & \textbf{lines} & \textbf{documents} \\
		\hline
	  iwslt\_2017\_ted      &  209522                  & 1705 \\
		\hline
		news-commentary-v18 &  449333                    & 11396 \\
		\hline
	\end{tabular}
	\caption{Document-level data used for LLM related experiments}
	\label{tab:doc-data}
\end{table}

\begin{table*}
	\centering
	\begin{tabular}{c}
		\hline
		\textbf{Streaming Translation Data Format} \\
		\hline
	  Natural English: <src1>, Translated German: <mt1>, Natural German:<ref1>   \\
        Natural German: <src2>, Translated English: <mt2>, Natural English:<ref2>  \\
        Natural English: <src3>, Translated German: <mt3>, Natural German:<ref3>  \\
        Translate the following sentence into German with a style bias towards Natural: \\  
        Natural English: <src4>, Translated German: <mt4>, Natural German: <ref4>  \\
		\hline
		\textbf{Context-aware Translation Data Format} \\
		\hline
        Natural English: <src1>, Translated German: <mt1>  \\
        Natural German: <src2>, Translated English: <mt2>  \\
        Natural English: <src3>, Translated German: <mt3>  \\
        Natural German: <src4>, Translated English: <mt4>  \\
        Natural English: <src5>, Translated German: <mt5>  \\
        Translate the following sentence into German with a style bias towards Natural: \\  
        Natural English: <src3>, Natural German: <ref3>  \\
        \hline
	\end{tabular}
	\caption{LLM Supervised fine-tuning(SFT) data format}
	\label{tab:data-format}
\end{table*}

\subsection{Data pre-processing}
Since the domain-specific data listed in Table~\ref{tab:bi-data} is limited, no special treatment was applied to this portion of the data; it was simply tokenized and input into the NMT model. For the document data in Table~\ref{tab:doc-data}, we constructed the two formats shown in Table~\ref{tab:data-format} by considering the characteristics of chat tasks, and used them to fine-tune the LLM, separately validating the impact of only preceding information and both preceding and context information on chat translation quality.

In the format of streamlined translation, during each translation session, only preceding information is visible. The LLM generates results based on this preceding information and the previews translation output, resulting in a translation that leans more towards the style of the reference.

In the context-aware translation format, during each translation session, preceding and following N sentences are provided along with the output of the NMT model, guiding the LLM to combine context information to produce a more natural translation.

\section{System Overview}
\label{sec:system}

\subsection{Model}

The baseline models for WMT24 chat task use the Transformer-Big architecture. Deep transformer is an improvement of Transformer, which increases the number of encoder layers and uses pre-layer-normalization to further improve model performance. Therefore, in this task, we adopt the following model architecture:
\begin{itemize}
	\item Deep 25-6 large Model: This model features 25-layer encoder, 6-layer decoder, 1024 dimensions of word vector, 4096 domensions of FFN, 16-head self-attention, and pre-layer-normalization.
\end{itemize}

For experiments related to large language model, we choose llama2-8b as the base.

\subsection{MBR Decoding}
Minimum Bayesian Risk (MBR) decoding was initially introduced during the era of statistical machine translation\citep{kumar-byrne-2004-minimum, pmlr-v235-jinnai24a}. This strategy calculates the output with the minimum expected error among multiple candidates, rather than simply selecting the result with the highest probability during the decoding process. In our experimental approach, we utilize the outputs of 10 distinct models as candidates. These candidates are then used to score each other’s comet, and the candidate with the highest average comet is chosen as the final output. Algorithm ~\ref{alg:MBR} show the detail.

\begin{algorithm}
\caption{MBR decoding algorithm}
\label{alg:MBR}
\begin{algorithmic}[1]
\REQUIRE ~~\\
    The set of translation candidates file, $MT_n$;\\
    The source text file, $SRC$;\\
    Comet metric model, $M_{comet}$;
\ENSURE final translation output ~~\\
    \STATE initialize output list $out[]$
    \FOR{each $line \in [MT_1, ... ,MT_n, SRC]$}
	    \STATE initialize $tmp\_max\_comet = 0$
		\STATE initialize $candidate\_mt = ''$
		\FOR{each $candidate \in [mt_1, mt_2, ... ,mt_n]$}
            \STATE let each $mt_x$ as ref, $candidate$ as mt and calculate the comet score with source text using $M_{comet}$
		  \STATE $mean\_comet = \frac{\sum_{x=1}^n comet_x}{n}$
		  \IF{$mean\_comet > tmp\_max\_comet$}
			\STATE $tmp\_max\_comet = mean\_comet$
				\STATE $candidate\_mt = candidate$
            \ENDIF
		\ENDFOR
		\STATE out.append($candidate\_mt$)
    \ENDFOR
\RETURN out
\end{algorithmic}
\end{algorithm}

\subsection{Regularized Dropout}
Regularized Dropout (R-Drop) \footnote{https://github.com/dropreg/R-Drop}\citep{2021arXiv210614448L} presents a simple yet more effective approach to regulate the training inconsistency caused by dropout \citep{JMLR:v15:srivastava14a}. Specifically, during each mini-batch training, each data sample is processed twice through the forward pass, with each pass utilizing a distinct sub-model and randomly dropping out some hidden units. R-Drop minimizes the bidirectional Kullback-Leibler (KL) divergence \citep{6832827} between the two distributions outputted by the two sub-models for the same data sample, thereby regulating the outputs of two sub-models randomly sampled from dropout for each data sample in training. This method effectively alleviates the inconsistency between the training and inference stages.

\subsection{Self-Training}
Self-Training(ST) \citep{imamura2018nict}, also known as forward translation (FT) \citep{wu2019exploiting}, typically involves utilizing a forward NMT model to translate source-side monolingual data into target-side text, thereby generating synthetic bilingual data. The generated data is then employed to train the forward translation model. Typically, beam search \citep{freitag2017beam} is applied for forward translation. In our experimental approach, we set the beam size to 4. Furthermore, we utilized the MBR selection results as self-training data, which led to the best results on the validation set.

\subsection{Back Translation}
Back-translation \citep{edunov2018understanding, wei2023textstyletransferbacktranslation} is acknowledged as a highly effective data augmentation strategy to boost NMT model performance. Unlike forward translation, back-translation converts target-side monolinguals into source-side text, thereby producing synthetic parallel corpora. Numerous back-translation techniques have been explored, with sampling \citep{gracca2019generalizing}, noise \citep{edunov2018understanding}, and tagged back-translation \citep{caswell2019tagged} demonstrating superior results. In our experimental setup, we opted for sampling back-translation.

\subsection{Model Averaging}
Model averaging \citep{https://doi.org/10.1002/ecm.1309} is a widely utilized technique to enhance translation quality. Typically, models (in our experiment, 5 models) that exhibit the highest performance on the development set are chosen for parameter averaging, which leads to substantial improvements.

\subsection{LLM Few-shot Prompting}
Although large language models exhibit impressive zero-shot capabilities, they still struggle with more complex tasks in the zero-shot setting. To address this, few-shot prompting can be employed as a technique for in-context learning, where demonstrations are provided in the prompt to guide the model towards enhanced performance. In our approach, we provide 5 reference translations to assist the large language model in producing superior results.

\begin{table*}[ht]
	\centering
	\begin{tabular}{lcccc}
		\hline
		\textbf{team} & \textbf{comet}$\uparrow$ & \textbf{chrf}$\uparrow$ & \textbf{bleu}$\uparrow$ & \textbf{context-comet-qe}$\uparrow$ \\
		\hline
		HW-TSC         & \textbf{93.4} & \textbf{83.2} & \textbf{69.8} & 0.221 \\
		unbabel+it     & 92.9          & 78.2          & 62            & \textbf{0.253} \\
		clteam         & 91.3          & 71.9          & 53            & 0.204 \\
		ADAPT          & 90.8          & 72.1          & 55            & 0.168 \\
        DCUGenNLP      & 90.8          & 71.2          & 53            & 0.188 \\
        baseline       & 89.8          & 70.8          & 51.1          & 0.173 \\
        SheffieldGate  & 89.4          & 67.5          & 45.2          & 0.177 \\
		\hline
	\end{tabular}
	\caption{The official automatic evaluation results of the test set, ranked based on the COMET-22 score}
	\label{tab:main_result}
\end{table*}

\subsection{LLM SFT with LoRA}
LLM SFT (Supervised Fine-Tuning) is a technique for fine-tuning large language models using specific datasets, which effectively enhances the performance of large language models on tasks such as text generation, machine translation, or sentiment analysis. LoRA (Low-Rank Adaptation)\citep{hu2022lora} is a technique that reduces the computational burden during large language model training by decreasing the number of model parameters through matrix decomposition. This technique maintains performance while lowering computational and memory requirements. By applying LoRA, large language models can perform better under limited computational resources, reducing training costs and resource consumption.

\section{Experiment Setting}
\label{sec:setting}
During the NMT model training phase, we use Pytorch-based Fairseq\footnote{\url{https://github.com/facebookresearch/fairseq}} \citep{ott2019fairseq} open-source framework as our benchmark system. Each model is trained using 8 GPUs with a batch size of 2048. The update frequency is 4 and the learning rate is 5e-4. The label smoothing rate is set to 0.1, the warm-up steps to 4000, and the dropout to 0.3. Adam optimizer \citep{kingma2015adam} with $\beta$1=0.9 and $\beta$2=0.98 is also used. Beyond that, we have configured the hyper parameter reg-alpha of the R-Drop technique to a value of 5. In the evaluation phase, We employ the official automatic evaluation scripts and primarily base our model and result selection on the comet metric\citep{rei-etal-2022-comet}\footnote{\url{https://github.com/Unbabel/COMET}}.

In the experiments related to large models, we utilize the open-source model llama2\_8b\_instruct from Meta and the training scripts from HF to train our models, setting the max\_seq\_length to 1024. For inference on large models, we employ the vllm tool.

\section{Result and Analysis}
\label{sec:result}

Table~\ref{tab:main_result} displays the results of the official test set, ranked according to the comet-22 score, where our system achieved the top position in comet-22, chrF, and BLEU metrics.

The primary results we submit are obtained by translating the source text of the test set with multiple NMT models, selecting the optimal output using MBR strategy, then training on the best models from the validation set using self-training method. The models are averaged over 5 epochs before being used to translate the test set to yield the final results.

\begin{table}
	\centering
	\begin{tabular}{lcc}
		\hline
		\textbf{System} & \textbf{en$\to$de} & \textbf{de$\to$en} \\
		\hline
		baseline              & 86.76 & 85.88 \\
        \hline
        22\_denoise           & 90.06 & 91.42 \\
		\hspace{1em}+ ST      & 91.23 & 91.40 \\
		\hspace{1em}+ ST\&BT  & 91.23 & 91.53 \\
		\hspace{2em}+ MBR ST  & \textbf{91.91} & \textbf{91.86} \\
        \hline
        MBR                   & 91.75 & 90.87 \\ 
		\hline
	\end{tabular}
	\caption{Sentence-level NMT results.}
	\label{tab:nmt_result}
\end{table}

\begin{table*}[ht]
	\centering
	\begin{tabular}{lcccc|cccc}
		\textbf  & \multicolumn{2}{c}{chat en$\to$de} & \multicolumn{2}{c|}{chat de$\to$en} & \multicolumn{2}{|c}{iwslt en$\to$de} & \multicolumn{2}{c}{iwslt de$\to$en} \\
        \textbf{System}      & comet & d-comet & comet & d-comet & comet & d-comet & comet & d-comet \\
		\hline
		Baseline             & 86.76 & 79.40 & 85.88 & 79.77 & - & - & - & - \\
		\hline
		MBR ST               & \textbf{91.91} & \textbf{85.41} & \textbf{91.86} & \textbf{86.21} & 84.70 & 77.55 & 87.05 & 80.81 \\
		\hline
		llama2\_8b\_instruce & 87.56 & 79.99 & 86.96 & 80.89 & 82.53 & 75.07 & 86.21 & 79.74 \\
        \hspace{1em}+ 5 best & 90.05 & 83.34 & 88.72 & 83.11 & 85.10 & 77.98 & 87.20 & 81.04 \\
		\hline
        stream               & 85.47 & 78.50 & 83.98 & 78.80 & \textbf{85.69} & \textbf{78.91} & \textbf{87.45} & \textbf{81.73} \\
        context-aware        & 81.82 & 73.81 & 83.89 & 77.37 & 84.80 & 77.51 & 86.65 & 80.43 \\
	\end{tabular}
	\caption{The results of LLM MT}
	\label{tab:llm_exp_res}
\end{table*}

\subsection{Sentence-level NMT}
In the previous chat tasks, we have tried various strategies to optimize the model, and the results from the validation set indicate that the baseline model from 2022 was already sufficiently powerful. On this basis, we combined this year’s training set, the 2022 validation and test sets, and conducted BT and ST reinforcement strategies, only in the direction of translation from English to German has there been a noticeable improvement. The results shown in Table~\ref{tab:nmt_result}.

To further improve the results, we attempted the MBR decoding strategy, generating 10 alternative outputs for the validation set using different NMT models in previous steps. These outputs were scored using comet, and the output with the lowest Bayesian risk was selected as the final result. The results in Table~\ref{tab:nmt_result} indicate that improvement was only seen in the en$\to$de direction. Further, we utilized the MBR results to perform another ST on each direction, ultimately achieving the best results in both directions in the validation set. The reason for the improvement we observed is that the MBR algorithm can integrate the capabilities of multiple models. When performing self training, it essentially utilizes the optimal results of multiple models for a round of knowledge distillation.

\subsection{Document-level MT with LLM}

According to the test results shown in Table~\ref{tab:llm_exp_res}, on the chat task valid set, the results of LLM (Large Language Model) are significantly worse than sentence-level under both comet or doc-comet metrics. The few-shot capabilities of LLM is indeed far better than zero-shot, but it still falls short of sentence-level results. After using the document-level data for LLM SFT, the results became even worse. We analyzed that the reason is the large domain shift, as the IWSLT and WMT datasets we used are far from the domain of the chat task.

To validate the capability of LLM in translating document-level content, we tested the results on the iwslt2017 en-de document-level test set. The results in the right half of Table~\ref{tab:llm_exp_res} demonstrate that LLM’s few-shot capability surpassed that of the chat task’s sentence-level model on this test set. Further, by fine-tuning the large model with document-level data, we obtained better results.

Comparing the results of stream translation and context-aware translation, we originally expected context-aware format data to yield better results because the model could refer to contextual information during translation. However, we analyzed that stream translation sees the previous step’s translation result each time, which is more consistent with the translation style of large model. On the contrary, context-aware requires input of the reference MT result from sentence-level model in one go, which is less consistent with the style of large model, causing the model to fail to effectively utilize these information.

\section{Conclusion}
\label{sec:conclusion}
This paper presents the submissions of HW-TSC to the WMT 2024 Chat Translation Shared Task. For both direction in en$\leftrightarrow$de translation task, we perform experiments with a series of training strategies. The results show that MBR self-training achieves the best results. In the future, we will continue to explore the applicability of MBR strategy mentioned in this paper. 

Beyond that, due to time constraints, further fine-tuning of large language models using chat task data was not conducted to assess its performance. Additionally, there is room for continued exploration of the translation capabilities of large language models.

\nocite{sutskever2014sequence, bahdanau2015neural, gehring2017convolutional, vaswani2017attention,yang-etal-2022-hw,Wang_Lyu_Ji_Zhang_Yu_Shi_Tu,dou2018exploiting,Moslem_Haque_Way_2023,Wu_Vu_Qu_Foster_Haffari_Anil_Dai_Firat_Lepikhin_Passos_et,kumar-byrne-2004-minimum,pmlr-v235-jinnai24a,2021arXiv210614448L,JMLR:v15:srivastava14a,6832827,imamura2018nict,wu2019exploiting,freitag2017beam,edunov2018understanding,gracca2019generalizing,caswell2019tagged,hu2022lora,ott2019fairseq,kingma2015adam,wei2023textstyletransferbacktranslation,wu-etal-2023-improving,wei-etal-2021-hw,guo2024novelparadigmboostingtranslation}

\bibliography{chat24}
\bibliographystyle{acl_natbib}

\end{document}